\documentclass[conference]{IEEEtran}

\usepackage{cite}
\ifCLASSINFOpdf
   \usepackage[pdftex]{graphicx}
\else
\fi
\usepackage{amsmath}
\usepackage{array}
\usepackage{amsfonts}
\usepackage{mdwmath}
\usepackage{amssymb}
\usepackage{mdwtab}
\usepackage{fixltx2e}
\usepackage{mathtools}

\begin{document}
\title{Energy Disaggregation for Real-Time Building Flexibility Detection}
\author{\IEEEauthorblockN{Elena Mocanu, Phuong H. Nguyen, Madeleine Gibescu}
\IEEEauthorblockA{Department of Electrical Engineering, Eindhoven University of Technology\\
5600 MB Eindhoven, The Netherlands\\
Email: \{e.mocanu, p.nguyen.hong, m.gibescu\}@tue.nl}
}

\maketitle

\begin{abstract}
Energy is a limited resource which has to be managed wisely, taking into account both supply-demand matching and capacity constraints in the distribution grid. One aspect of the smart energy management at the building level is given by the problem of real-time detection of flexible demand available. In this paper we propose the use of energy disaggregation techniques to perform this task. Firstly, we investigate the use of existing classification methods to perform energy disaggregation. A comparison is performed between four classifiers, namely Naive Bayes, k-Nearest Neighbors, Support Vector Machine and AdaBoost.  Secondly, we propose the use of Restricted Boltzmann Machine to automatically perform feature extraction. The extracted features are then used as inputs to the four classifiers and consequently shown to improve their accuracy. The efficiency of our approach is demonstrated on a real database consisting of detailed appliance-level measurements with high temporal resolution, which has been used for energy disaggregation in previous studies, namely the REDD. The results show robustness and good generalization capabilities to newly presented buildings with at least 96\% accuracy.

\end{abstract}
\IEEEpeerreviewmaketitle

\section{Introduction}

Energy is a limited resource which faces additional challenges due to recent efficiency and de-carbonization goals worldwide. An important component of the ongoing process is the improvement in the energy management systems in residential and commercial buildings, which account for  $30-40\%$ of the total energy demand in the developed world \cite{Nejat2015843}. Buildings are complex systems composed by a different number of devices and appliances, such as refrigerators, microwaves, cooking stoves, washing machines etc. However, there are also a number of sub-systems, e.g. electric heating, lighting. Even there are many influencing factors in building energy consumption, some patterns can be clearly identified and used further to improve demand side management systems and demand response (DR) programs~\cite{6687966}. Identifying and aggregating the flexibility resource at the community level can decrease the end-user energy bill. Concomitantly, as a long-term benefit, flexibility can lead also to emission reductions, and lower investments in transmission and distribution grid infrastructure. Therefore, the role of end-users and their available flexibility is becoming increasingly important in the Smart Grid context. 
\footnote{\textbf{This article is a pre-print version.} Please cite this article as: E.~Mocanu, P.~H.~Nguyen and M.~Gibescu, \textit{Energy Disaggregation for Real-Time Building Flexibility Detection}, IEEE Power and Energy Society General Meeting, Boston, USA, 2016}

One possible way to detect building flexibility in real-time is by performing energy disaggregation. Disaggregation refers to the extraction of appliance level energy signals from an aggregate, or the whole-building, energy consumption signal. Often only this aggregated signal is made available via the smart meter infrastructure to the grid operator, due to privacy concerns of the end user. This new approach should open new paths towards better planning and operation of the smart grid, helping the transition of end-users from a passive to an active role. In addition, informing the end-user in real-time, or near real-time, about how much energy is used by each appliance can be a first step in voluntarily decreasing the overall energy consumption.

Introduced by W. Hart\cite{192069} in the early 1980s, the Non-Intrusive Load Monitoring (NILM) problem has nowadays several solutions for residential buildings. Traditional approaches for the energy disaggregation problem (or NILM problem) start by investigating if the device is turned on/off \cite{97667}, and followed by many steady-state methods \cite{1192027} and transient-state methods \cite{1192027} aiming to identify more complex appliance patterns. In the same time, advance building energy managements systems are looking beyond quantification of the energy consumption by including fusion information such as, the acoustic sensors to identify the operational state of the appliances \cite{Guvensan20131539}, the motion sensors, the frequency of the appliance used \cite{siam2011}, as well as time and appliance usage duration\cite{NIPS2010kolter, siam2011}. A more comprehensive discussion about these can be found in recent reviews, such as \cite{5618423,5735484,s121216838}. Moreover, new data analytics challenges arise in the context of an increasing number of smart meters, and consequently, a big volume of data, which highlights the need of more complex methods to analyze and take benefit of the fusion information \cite{6903193}. More recent researches have explored a wide range of different machine learnings methods, using both supervised and unsupervised learning, such us sparse coding \cite{NIPS2010kolter}, clustering \cite{5759180, 6939461} or different graphical models (e.g.  Factorial Hidden Markov models (FHMM)\cite{siam2011}, Factorial Hidden Semi-Markov Model (FHSMM) \cite{siam2011}, Conditional FHMM \cite{siam2011}, Conditional Factorial Hidden Semi-Markov Model (CFHSMM)\cite{siam2011}, additive FHMM \cite{AISTATS2012_KolterJ12} or Bayesian Nonparametric Hidden Semi-Markov Models \cite{Johnson}) to perform energy disaggregation. Still, there is an evident challenge to develop an accurate solution that could perform well for every type of appliance.

In this paper, the aim is to perform real-time flexibility detection  using  energy disaggregation techniques. Therefore, the key methodological contribution of this paper is a machine learning based tool for exploiting the building energy disaggregation capabilities in an online manner. Our contributions can be summarized as follows. Firstly, we investigate the use of classification methods to perform energy disaggregation. Consequently, a comparison is performed between four widely-used classification methods, namely Naive Bayes (NB), K-Nearest Neighbors (KNN), Support Vector Machine (SVM) and AdaBoost. Secondly, we introduce a Restricted Boltzmann Machine (RBM) to perform automatic feature extraction in order to improve the performance of the four classification methods discussed. We validate our proposed approach by using a real measurement database, specifically conceived for energy disaggregation, i.e. the REDD~\cite{kolter2011}. 

The remaining of this paper is organized as follows. Section~\ref{Sec:problem} introduces the problem description.  Section~\ref{Sec:methods} describes our proposed approach for the energy disaggregation problem. In Section \ref{Sec:Exp} the experimental validation of the proposed methods is detailed and Section~\ref{Sec:con} concludes the paper.

\section{Problem Formulation and Methodology}\label{Sec:problem}
This section details the problem definition targeted in this paper. In one unified framework, we split the problem into two parts, where first the energy disaggregation problem is solved, and then an identification procedure is carried out to analyze the potential of building demand flexibility.

The proposed solution for energy disaggregation is addressed using four different classification methods. More formally, let us define an input space  $\mathcal{D}$ and an output space (label space) $\mathcal{B}$.  The question of learning is reduced to the question of estimating a functional relationship of the form $\mathcal{C} : D \rightarrow B$, that is a relationship between inputs and outputs. A classification algorithm is a procedure that takes the training data as input and outputs a classifier $\mathcal{C}$. The goal is then to find a $\mathcal{C}$ which makes “as few errors as possible”. Intuitively, the learned classifier should be based on \textit{enough} training examples, \textit{fit} the training example and should be \textit{simple}. Moreover, classification can be thought of as two separate problems: binary classification and multi-class classification.
\begin{figure}[h!]
\centering
\includegraphics[scale=0.46]{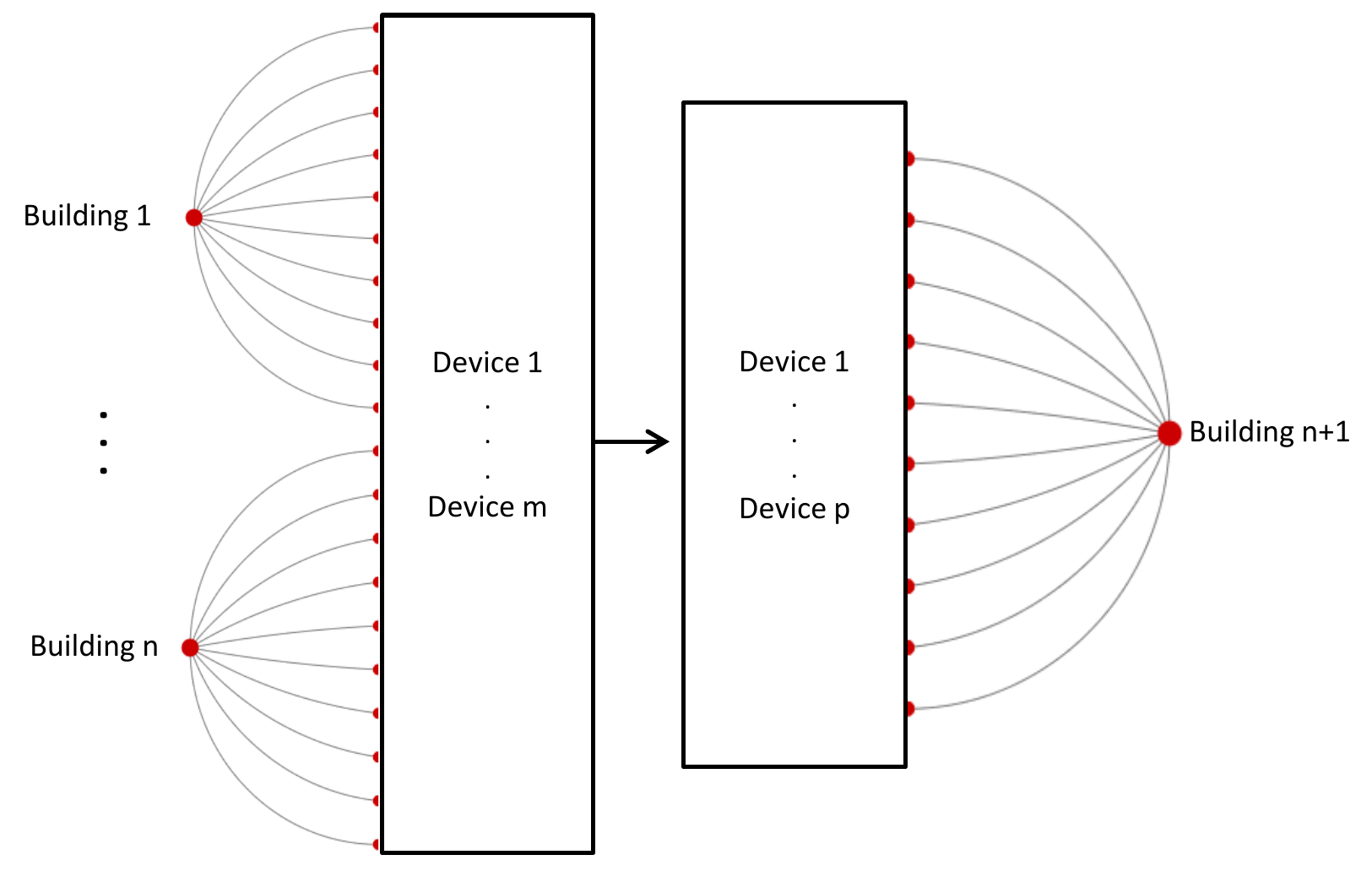}
\caption{Energy disaggregation}
\label{fig:class}
\end{figure}

In our specific case, the $B$ space is given by the electrical devices in the building, and the $D$ space is given by the aggregated electrical energy consumption of the building.  In Figure~\ref{fig:class}  the flow diagram of the energy disaggregation procedure is depicted. Firstly, using data from $n$ buildings we derive a corresponding model for each device inside them. Furthermore these binary classification models are used to automatically classify, whether a given device is active  at any specific moment in time, by using the building's total electrical energy consumption profile.

\section{Proposed Methods}\label{Sec:methods}
In this section, we  firstly  briefly describe the four classification methods to perform energy disaggregation, these methods being part of the supervised learning paradigm. Secondly, we introduce the mathematical details of the Restricted Boltzmann Machine used to perform automatic features extraction, this method being part of the unsupervised learning paradigm.

\subsection{Classification methods}
 For the classification problem, plenty of deterministic or probabilistic algorithms are known, where every observation is analyzed into a set of quantifiable properties, such as  Naive Bayes~\cite{bishopbook}, Support Vector Machine~\cite{vapniksvm}, AdaBoost~\cite{Freund99ashort}, Random Forest Trees and so on. Prior studies tried to determine the most accurate classification method, as is shown in\cite{CaruanaNiculescu}, but currently there is not a general consensus in the favor of a particular method.

\subsubsection{Naive Bayes} is one of the most simple classification method based on a strong independence assumptions between the input features. Despite these relatively naive assumptions, with a training phase extremely easy to implement and fast computational time, Naive Bayes classifiers often outperform more sophisticated alternatives. 
\subsubsection{k-Nearest Neighbors} is a non-parametric method used for classification.
The standard version of KNN used in this paper performs successively two steps. Specifically, the clusters are construct by partitioning the $k$-nearest neighbors based on a distance measure (i.e. Euclidean distance), followed by an update rule, such that the majority of those $k$-nearest neighbors decide the class of the next observations.
\subsubsection{AdaBoost}
it stands for Adaptive Boosting, and is a machine learning algorithm, which was proposed in the computational learning theory field by Y. Freund and R. Schapire~\cite{Freund99ashort}. AdaBoost method solves the classification problem using a linear combination of many weak classifiers into a single strong classifier.  Acting as an expert, boosting often does not suffer from overfitting and it is worth to investigate in the context of our challenging dataset.
\subsubsection{Support Vector Machine (SVM)}
is introduced by Vapnik in 1995~\cite{vapniksvm} and becomes very popular for solving problems in classification, regression, and novelty detection. An important characteristic of SVM is that the determination of the model parameters corresponds to a convex optimization problem, and so any local solution is also a global optimum. This guarantee comes with some computational cost but also with a better robustness. 
\subsection{Restricted Boltzmann Machine}\label{tab:rbm}
Restricted Boltzmann Machine is a two-layer generative stochastic neural network which is capable to learn a probability distribution over its set of inputs~\cite{originalrbm}. Such a model does not allow intra-layer connections between the units, and it allows just inter-layer connections. In fact, any unit from one layer has undirected connections to all the units from the other layers. Up to now, various types of restricted Boltzmann machines are already developed and successfully applied in different applications~\cite{eu}. Despite their differences, almost all of these architectures preserve RBMs characteristics. To formalize a restricted Boltzmann machine, and its variants, three main ingredients are required, namely an energy function providing scalar values for a given configuration of the network, the probabilistic inference and the learning rules required for fitting the free parameters.

Thus, a RBM consists in two binary layers, the visible  layer,  $\mathbf{v}=[v_1, v_2, .., v_{n_v}]$, in which each neuron represents one dimension (feature) of the input data and the hidden layer, $\mathbf{h}=[h_1, h_2, .., h_{n_h}]$, which represents hidden features extracted automatically by the RBM model from the input data, where $n_v$ is the number of visible neurons and $n_h$ is the number of the hidden neurons. Each visible neuron $i$ is connected to any hidden neuron $j$ by a weight, i.e. $W_{ij}$. All these weights are stored in a matrix $\mathbf{W}\in\mathcal{R}^{{n_v}\times{n_h}}$, where $\mathcal{R}$ is the set of real numbers, in which the rows represent the visible neurons and the columns the hidden ones. Finally, each visible neuron $i$ has associated a bias $a_i$ which is stored in a vector $\mathbf{a}=[a_1, a_2, .., a_{n_v}]$. Similarly, the hidden neurons have biases which are stored in a vector $\mathbf{b}=[b_1, b_2, .., b_{n_h}]$. Further on, we will note with $\mathbf{\Theta}=\{\mathbf{W},\mathbf{a},\mathbf{b}\}$ a set which represent the union of all free parameters of a RBM (i.e. weights and biases). Formally, the energy function of a RBM for any state $\{\mathbf{v},\mathbf{h}\}$ can be computed by summing over all possible interactions between neurons, weights and biases, as folows:
\begin{equation}
E(v,h)=-\sum_{i=1}^{n_v} \sum_{j=1}^{n_h} v_{i}h_{j}W_{ij}-\sum_{i=1}^{n_v} v_{i}a_{i}-\sum_{j=1}^{n_h}h_{j}b_{j}
 \label{Eq:EnergyRBM}
 \end{equation}
where the term $\sum\displaystyle_{i=1}^{n_v} \sum\displaystyle_{j=1}^{n_h} v_{i}h_{j}W_{ij}$ is given by the total energy between the neurons from different layers, while $\sum\displaystyle_{i=1}^{n_v}v_{i}a_{i}$ represents the energy of the visible neurons and $\sum\displaystyle_{j=1}^{n_h}h_{j}b_{j}$ is the energy of the hidden neurons. 
 
The inference in a RBM means to determine two conditional distributions. For any hidden or visible neuron this can be done just by sampling from a sigmoid function, as shown below:
\begin{align}
p(h_j=1|\mathbf{v},\mathbf{\Theta})=\frac{1}{1+e^{-(b_j+\sum\displaystyle_{i=1}^{n_v} v_{i}w_{ij})}}\\
p(v_i=1|\mathbf{h},\mathbf{\Theta})=\frac{1}{1+e^{-(a_i+\sum\displaystyle_{j=1}^{n_h} h_{j}w_{ij})}}
\label{Eq:XBMinferencevisible}
\end{align}
To learn the parameters of a RBM model there are more variants in the literature (e.g. persistent contrastive divergence, parallel tempering~\cite{desjardins:aistats2010}, fast persistent contrastive divergence~\cite{Tieleman:2009:UFW:1553374.1553506}). Almost all of them being derived from the Contrastive Divergence (CD) method proposed by Hinton in~\cite{hintoncd}. For this reason, in this paper, we briefly describe and use just the original CD method. CD is an approximation of the maximum likelihood learning, which is practically intractable in a RBM. Thus, while in maximum likelihood  the learning phase minimizes the Kullback-Leiber (KL) measure between the distribution of the input data and the model approximation, in CD the learning follows the gradient of: 
\begin{equation}
CD_{n} \propto D_{KL}(p_{0}(\textbf{x})||p_{\infty}(\textbf{x}))-D_{KL}(p_{n}(\textbf{x})||p_{\infty}(\textbf{x}))
\end{equation}
where, $p_{n}(.)$ represents the resulting distribution of a Markov chain running for $n$ steps. Furthermore, the general update rule of the free parameters of a RBM model is given by:
\begin{equation}
\Delta\mathbf{\Theta}_{\tau+1}=\rho\Delta\mathbf{\Theta}_{\tau}+\alpha(\nabla\mathbf{\Theta}_{\tau+1}-\xi\mathbf{\Theta}_{\tau})
\label{eq:genuprule}
\end{equation}
where $\tau$, $\alpha$, $\rho$, and $\xi$ represent the update number, learning rate, momentum, and weights decay, respectively, as thoroughly discussed in~\cite{hintontrain}. Moreover, $\nabla\mathbf{\Theta}_{\tau+1}$ for each free parameter may be computed by deriving the energy function from Equation~\ref{Eq:EnergyRBM} with respect to that parameter, as detailed in~\cite{hintoncd}, yielding:
\begin{align}
\nabla w_{ij}&\propto{\langle}v_i h_j{\rangle}_{0}-{\langle}v_i h_j{\rangle}_{n}\\
\nabla a_{i}&\propto{\langle}{v_i}{\rangle}_{0}-{\langle}v_i{\rangle}_{n}\\
\nabla b_{j}&\propto{\langle}{h_j}{\rangle}_{0}-{\langle}h_j{\rangle}_{n} 
\label{Eq:GXBMupdateb}
\end{align}
with $\langle\cdot\rangle_n$ being the distribution of the model obtained after $n$ steps of Gibbs sampling in a Markov Chain which starts from the original data distribution $\langle\cdot\rangle_0$.

\section{Experimental Results}\label{Sec:Exp}
In this section we analyze and validate our proposed approach using a real-world database, namely \textit{The Reference Energy Disaggregation Dataset} (REDD), described by Kolter and Johnson in~\cite{kolter2011}. This data was chosen as it is an open dataset\footnote{http://redd.csail.mit.edu/, Last visit November 5th, 2015} collected specifically for evaluating energy disaggregation methods. It contains aggregated data recorded from six buildings over few weeks sampled at 1 second resolution, together with the specific data for all appliances of each building at 3 seconds resolution.

In the first set of experiments, we study the performance of the classification methods (i.e. Naive Bayes, K-Nearest Neighbors, Support Vector Machine and AdaBoost) for detecting the activation of four appliances (i.e. refrigerator, electric heater, washer-dryer, dishwasher), specifically chosen for their ability to provide demand-side flexibility. Furthermore, in the second stage we demonstrate the improvement in the accuracy of the classification after a Restricted Boltzmann Machine is used for automatic feature extraction. Finally, assuming the aforementioned four appliances shiftable in time, we discuss the possible benefits of real-time flexibility detection.

The experiments were performed in the MATLAB\textsuperscript{\textregistered} environment using the methods described in Section~\ref{Sec:methods}. For the classification methods we have used the optimized parameters from the  machine learning toolbox (e.g. SVM with radial kernel function). For each appliance we have built a separate binary classification model for every classification method. The input at every moment in time is given by a window of 10 consecutive time steps from the aggregated building consumption, while the output was represented by the activation of the appliance (i.e. on/off status). In all the experiments performed, we have trained the models on 5 buildings (i.e. 2, 3, 4, 5, and 6) and we have tested the models on a different building (i.e. 1). Also, as recommended in~\cite{6939461}, we have applied a median filter of 6 samples to make the data smoother.

 For the feature extraction procedure we have implemented RBMs with the following parameters: 20 hidden neurons and 10 visible neurons (representing the time window of 10 consecutive time steps).  After a short fine tuning procedure, the learning rate was set to $10^{-2}$, the momentum was set to 0.5, and the weight decay was set to 0.0002. We trained the RBM models for 25 epochs, and after that we have used the probabilities of the hidden neurons as inputs for the classification methods.

In order to characterize as fairly as possible the accuracy of the models proposed to classify the appliance activation we have calculated the classifier accuracy as follows:
\begin{equation}
Accuracy=\frac{\sum_{i=1}^{n} A_{ii} }{\sum_{i=1}^{n}\sum_{j=1}^{n} A_{ij}}
\end{equation}
where $A$ is the confusion matrix (also known as a contingency table or an error matrix), $A_{ii}$ represents the positive true value and the denominator represents the total number of data used in the classification procedure. This quantifies the proportion of the total number of instances that were correctly classified.

\subsection{Energy disaggregation}
In this subsection, we first perform a comparison between the four classification methods, namely Naive Bayes (NB), k-Nearest Neighbors (KNN), Support Vector Machine (SVM) and AdaBoost (AB). Table~\ref{tab:2class} summarizes the classification accuracy for different building electrical components, such as refrigerator, electric heater, washer-dryer and dishwasher.   For a better insight into the results, an example of the energy consumption for the appliances corresponding to building 1 (the test data) is depicted in Figure~\ref{fig:devices}.
\begin{table}[ht!]
\caption{ Results showing accuracy [\%] for each of  Naive Bayes, KNN, SVM and AdaBoost to classify an appliance versus all data.}
\centering 
\begin{tabular}{|c|cccc|}
\hline
Appliance&NB&KNN&SVM&AdaBoost\\
\hline
refrigerator&52.18\%&67.36\%&67.45\%&\textbf{87.13\%}\\
electric heater&93.01\%&97.79\%&\textbf{98.84\%}&94.74\%\\
washer dryer&92.04\%&\textbf{96.17\%}&78.27\%&95.56\%\\
dishwasher&97.52\%&\textbf{98.11\%}&97.74\%&97.77\%\\
\hline
\end{tabular}
\label{tab:2class}
\end{table}
\begin{figure}[h!]
\centering
\includegraphics[scale=0.63]{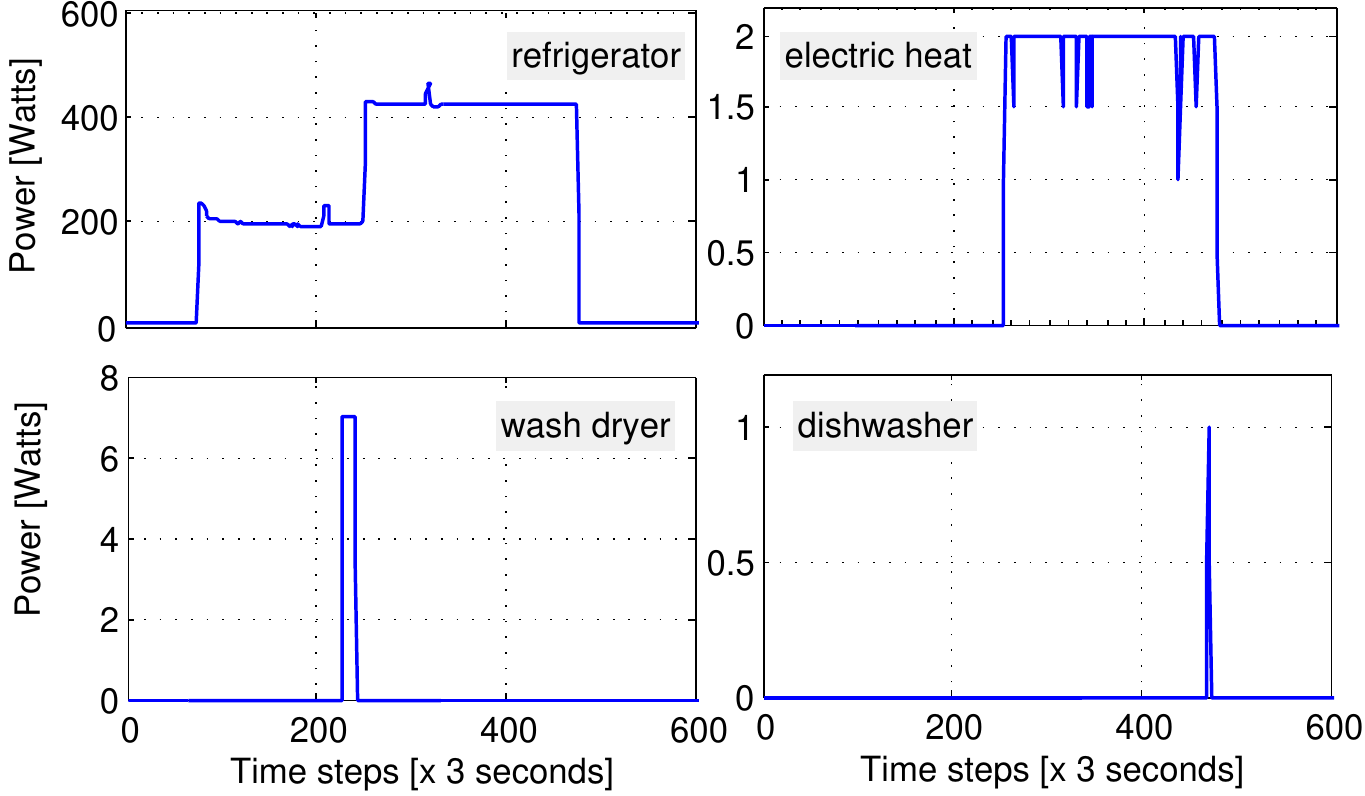}
\caption{An example of energy consumption in Building 1 over 30 minutes for refrigerator, electric heater, washer dryer and dishwasher.}
\label{fig:devices}
\end{figure}

Furthermore, to improve the classification performance, we have employed the automatic features extraction procedure by using the Restricted Boltzmann Machine as described in Section\ref{tab:rbm}. Next, the extracted features  are used as inputs for the classification methods.
We have tested and validated this approach on the same electrical appliances as before, as shown in Table~\ref{tab:2classrbm}.
\begin{table}[ht!]
\caption{ Results showing accuracy [\%] for each of Na$\ddot{\i}$ve Base, KNN, SVM and AdaBoost with RBM extension, to classify an appliance versus all data.}
\centering 
\begin{tabular}{|c|cccc|}
\hline
Appliance&NB-RBM&KNN-RBM&SVM-RBM&AB-RBM\\
\hline
refrigerator&64.78\%&\textbf{96.72\%}&84.45\%&91.02\%\\
electric heater&99.13\%&99.81\%&\textbf{99.86}\%&99.84\%\\
washer dryer&99.14\%&97.31\%&89.23\%&\textbf{99.27}\%\\
dishwasher&97.64\%&98.43\%&\textbf{98.67}\%&97.82\% \\
\hline
\end{tabular}
\label{tab:2classrbm}
\end{table}
It can be observed that in all situations, the use of RBMs has improved the accuracy for each classifier. This culminates with an improvement of around 30\% for the case of the refrigerator classified with KNN, from 67.36\% initial accuracy, up to 96.72\% accuracy after the use of RBM. It is worth mentioning, that the imbalanced number of data points in every class  suggests that a more deeper data mining analysis may be useful. In term of computational complexity the training time varies from the range of few seconds in the case of KNN up to few minutes in the case of SVM. In the testing phase, to classify all the data points considered (i.e. 745868 instances per year per appliance) each of the methods has ran in approximately 1 second, except SVM which ran in 4-5 seconds. Overall, this yields an execution time of a few microseconds per data point making the approach suitable for a large range of real-time applications.

\subsection{Flexibility detection}
The energy disaggregation results may be used further in a large number of applications, as reported in 2015 by the US Department of Energy in an extensive report \cite{Report} which aims to characterize the actual performance of energy disaggregation solutions used in both the academic research and in commercial products.

Most importantly, our results may be used to detect in real-time the building flexibility available. We observed that approximately 17\% of the total energy consumption for building 1 is used by the four disaggregated appliances, such as  refrigerator 11.72\%, electric heater 5.08\%, washer-dryer 0.0007\% and dishwasher 0.9\% respectively. More statistical details about these appliances for building 1 are presented in Table~\ref{fig:devices}.
\begin{table}[ht!]\label{tab:summary}
\footnotesize
\centering
\caption{General characteristics of the building 1 appliances used in the experiments.} 
\begin{tabular}{ccc}
&Mean&Standard deviation\\
\hline
\hline
refrigerator&56.41&86.65\\
electric heater&24.44&148.16\\
wash dryer&0.11&0.96\\
dishwasher &4.30&43.54\\
\hline
\end{tabular}
\label{tab:summary}
\vspace{-3mm}
\end{table}
A visual examination of the results, assuming that all the four appliances studied have smart time-shifting capabilities, and a detection accuracy of over 96\%  in all the experiments, show a significant peak reduction. As by example, in Figure~\ref{fig:devices} the inflexible load is represented by the difference between the total energy consumption signal and the sum of our disaggregated signals over 24  hours. In this case, we may observe that the average buildings flexibility is 23.21\%.
\vspace{-3mm}
\begin{figure}[h!]
\centering
\includegraphics[scale=0.61]{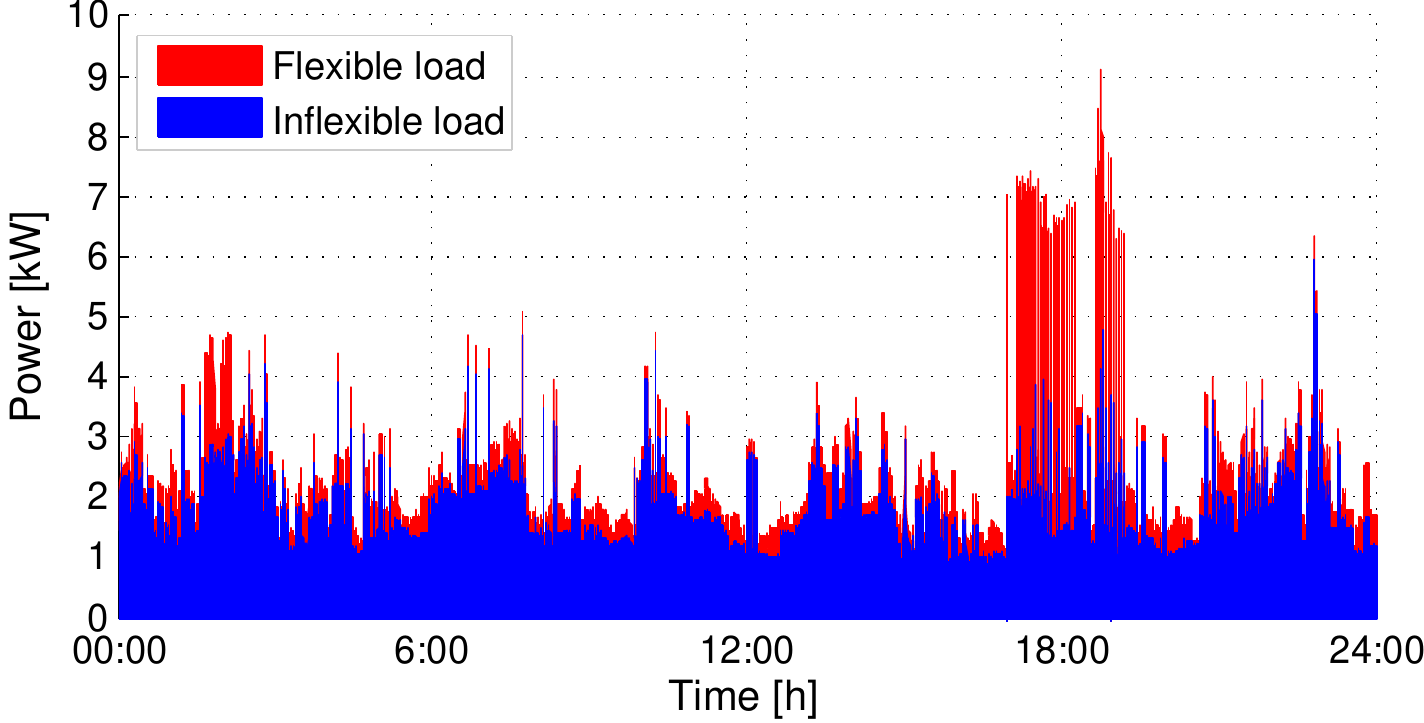}
\caption{An example of electrical energy consumption in buildings over one day for inflexible load and flexible load (refrigerator, electric heater, washer dryer and dishwasher).}
\label{fig:devices}
\vspace{-5mm}
\end{figure}
\section{Conclusion}\label{Sec:con}
In this paper a novel tool capable to perform accurate energy disaggregation for real-time flexibility detection is proposed. A comparison between four existing classification methods was performed. Aiming at enhancing the quality of such estimates as well as at increasing the accuracy of energy disaggregation, a method for automatic features extraction is proposed, using Restricted Boltzmann Machines. By incorporating the RBM for feature extraction, each of the classification methods, i.e. Naive Bayes, k-Nearest Neighbors, Support Vector Machine and AdaBoost, has outperformed its non-preprocessed counterpart.  The experimental validation performed on the REDD dataset shows that KNN- RBM has the best trade-off between accuracy and speed.

\section*{Acknowledgment}
This research has been funded by NL Enterprise Agency under the TKI Switch2SmartGrids project of Dutch Top Sector Energy.

\vspace{-1mm}
\bibliographystyle{IEEEtran}
\bibliography{mybib}

\end{document}